\documentclass[10pt,twocolumn,letterpaper]{article}

\usepackage{cvpr}
\usepackage{times}
\usepackage{epsfig}
\usepackage{graphicx}
\usepackage{amsmath}
\usepackage{amssymb}
\usepackage{bm}
\usepackage{multirow}
\usepackage{array}

\usepackage[breaklinks=true,bookmarks=false]{hyperref}

\cvprfinalcopy 

\def\cvprPaperID{****} 

\begin{document}

\title{On Study of the Binarized Deep Neural Network for Image Classification}

\author{Song Wang, Dongchun Ren, Li Chen, Wei Fan, Jun Sun, Satoshi Naoi\\
{Fujitsu Research \& Development Center, Beijing, China}\\
\{wang.song, rendongchun, chenli, fanwei, sunjun, naoi\}@cn.fujitsu.com\\
}

\maketitle

\begin{abstract}
   Recently, the deep neural network (derived from the artificial neural network) has attracted many researchers' attention by its outstanding performance. However, since this network requires high-performance GPUs and large storage, it is very hard to use it on individual devices. In order to improve the deep neural network, many trials have been made by refining the network structure or training strategy. Unlike those trials, in this paper, we focused on the basic propagation function of the artificial neural network and proposed the binarized deep neural network. This network is a pure binary system, in which all the values and calculations are binarized. As a result, our network can save a lot of computational resource and storage. Therefore, it is possible to use it on various devices. Moreover, the experimental results proved the feasibility of the proposed network.    
\end{abstract}

\numberwithin{equation}{section}
\section{Introduction} \label{introduction}

\begin{figure}[h]
\begin{center}
\includegraphics[width=0.3\textwidth]{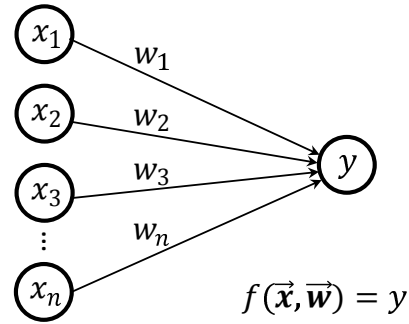}
\caption{The basic function of ANN. \label{ANNbasic}}
\end{center}
\end{figure}

The research of artificial neural networks (ANN) began more than 70 years ago, proposed by Warren McCulloch and Walter Pitts~\cite{Pitts}, Donald Hebb~\cite{Hebb} and Frank Rosenblatt~\cite{Rosenblatt}. Especially in~\cite{Rosenblatt}, a two-layer network is introduced for pattern recognition. Figure~\ref{ANNbasic} shows the basic function of ANN, that is, the neuron of higher layer is calculated by the neurons of prior layer with the connecting weights. However, there was no solution for the network training until backpropagation (gradient descent) algorithm was created by Paul Werbos~\cite{Werbos}. After James McClelland~\cite{McClelland} introduced the ANN as simulation of natural neural process and its usage in artificial intelligence (AI), the research of ANN became popular. Since then, ANN was successfully applied to image classification~\cite{Benediktsson, Bischof, Giacinto}, character recognition ~\cite{Rajavelu}, face recognition~\cite{Lawrence}, speech recognition~\cite{Waibel} and so on. Moreover, a theoretical explanation of ANN's success was also given by Kurt Hornik~\cite{Hornik}. \par

Because of the limitation of computers, the ANN research stagnated until the new century came. With the development of computers, the training of large-scale neural network became possible. As a result, the research of deep neural networks (DNN) emerged and attracted more and more attention. For example, Geoff Hinton's deep belief nets~\cite{Hinton}, Yann LeCun and Dan Ciresan's study on deep convolutional neural networks~\cite{LeCun, Ciresan}, and Alex Graves's deep recurrent neural networks~\cite{Graves}. The features of DNN can be summarized as follows.
\begin{itemize}
\item Large-scale network: the DNN model usually has very large structure (with many layers), including millions of neurons and connections. Consequently, the computational cost of DNN is extremely high, thus most of the DNN experiments are done with GPUs.
\item Better performance: DNN archived the state-of-the-art results on various tasks and competitions, and brought breakthroughs to those fields, such as handwritten character recognition~\cite{Graves3, Graves4}, image classification~\cite{Ciresan, Ciresan2}, speech recognition~\cite{Graves2} and so on.
\item Lack of theoretical explanation: although the DNN achieved much better performance than conventional methods, there is no convincing explanation for such success. For example, it is well known that some trick in training could improve the performance of DNN significantly, but it was hard to explain such effect theoretically. 
\end{itemize}
\par

Obviously, although the performance of DNN is promising, it still needs to be improved in different ways. On one hand, in order to pursue higher recognition rate, several optimization methods for training were proposed, such as dropout~\cite{dropout} and dropconnect~\cite{dropconnect}. Those methods can reduce the overfitting problem significantly. On the other hand, new understanding of the neural network emerges and may extend the ability of DNN. For example, in Ian Goodfellow's recent work~\cite{Goodfellow} for digit string recognition, the output layer is trained to show both the digit number and the recognition result of each digit. By doing this, their DNN model is able to recognize the digit string directly, without any segmentation process. This work extended the DNN from single character recognition to character string recognition. Inspired by this work, we may change the basic framework of DNN to find more possibilities.\par

In this paper, we focus on the basic function of ANN and try to make it more suitable for computers. Usually, the ANN is seen as simulation of natural neural process, thus its neurons and weights are all real number. Such values can represent the electric signal generated by neural cells. However, the ANN models are often realized by computers. As we all know, the computer process data based on binary value, like ``0'' and ``1''. In other words, we can say that the basic ``neural cell'' of computer generates binary signals. Inspired by this observation, we proposed a new type of neural network --- the binarized deep neural network (BDNN). In BDNN, all the neurons and weights are binary value; at the same time, the calculation of the basic function of BDNN is also Boolean. \par

Actually, there were researches of binary neural network (Hopfield neural network)~\cite{Hopfield, Hopfield2, Hopfield3, Muselli, Takefuji} and corresponding training algorithms~\cite{Gray, Kim}. Nevertheless, this kind of neural network is quite different from BDNN. Although the input and output of the binary neural network are binary values, the weights of which are real number. Therefore, the calculation of binary neural network is not different from the conventional neural network. In contrast, the BDNN is a pure binary system, in which all the variables and operations are all binarized (Boolean). 

Compared with conventional deep neural network, the BDNN is expected to have several promising merits, which are shown as follows. 
\begin{itemize}
\item Less storage request: since all the weights in BDNN are binary value, thus we can use only 1 bit to store one weight. As a result, we can save a large network with small storage. In contrast, we must use at least 16 bits to store one real number weight of conventional neural network.
\item Higher speed on CPU: the basic calculation of CPU is the Boolean calculation. For CPU, this calculation is the most efficient. Since the BDNN only uses Boolean calculation, thus the processing speed of which can be easily optimized on CPU. Consequently, it is possible to run BDNN of high speed on devices with only CPUs. In contrast, conventional DNN can only run fast with GPUs. 


\item Clear network response: in BDNN, it is quite easy to observe the response of neurons and weights in the propagation process. This is because each neuron or weight only has two different kinds of status. This may be helpful for us to design proper learning strategy for BDNN. 
\end{itemize}
\par

In summary, with binary variables and Boolean operations, the BDNN is able to run with reasonable computational resource and storage. Although now the performance of BDNN may not be comparable with the conventional deep neural network of the same scale, it has the potential to be improved in the future. \par

This paper is organized as follows. In Section~\ref{BDNN}, the principles of the BDNN is introduced as well as the hybrid binarized deep neural network (hybrid-BDNN) for non-binary input data. In Section~\ref{training}, a training method is proposed for BDNN and hybrid-BDNN. In Section~\ref{experiments}, comparison experiments of BDNN and conventional DNN are analyzed. The last section is the conclusion part. \par

\section{The binarized deep neural network} \label{BDNN}

In this section, the recognition process (forward propagation) of BDNN is introduced. As mentioned above, only binary variables and Boolean operations are used in this process. Moreover, in order to use BDNN on non-binary input data, we also introduced the hybrid-BDNN, which contained both conventional neural network part and BDNN part. \par

\subsection{The basic function for BDNN}
As shown in Fig.~\ref{ANNbasic}, with this basic function, we can build network of any complicated structure. Therefore, the BDNN is actually a new definition of the basic function. Just like the conventional DNN, with the basic function of BDNN, any neural network structure can be built.\par

Usually, a basic function of neural network should contain two different types of calculation: the linear and nonlinear calculations. For example, as shown in Fig.~\ref{ANNbasic}, in conventional neural network, the linear calculation is an inner product of the input neurons $\bm{\vec x}$ and the corresponding weights $\bm{\vec w}$, that is, the $\bm{\vec x} \cdot \bm{\vec w}$; the nonlinear calculation of conventional neural network is activation function (sigmoid, hyperbolic tangent, etc.) or pooling (often used in DNN). Consequently, in the definition of the basic function of BDNN, we also defined both the linear and nonlinear calculations. \par

First, since the BDNN is created by binary values, the basic calculation should be chosen from Boolean algebra. The Boolean binary operations are ``and $\wedge$'', ``or $\vee$'' and several derived operations (``exclusive or $\oplus$'', ``equivalence $\equiv$'', ``material implication $\rightarrow$''). The truth table of those operations is shown in Table~\ref{truthtable}. Obviously, the ``exclusive or'' and ``equivalence'' should be chosen because their results are balanced between ``0'' and ``1''. In BDNN, we chose the ``equivalence'' as the linear calculation. If we use $-1$ to represent ``0'' and $1$ to represent ``1'', then the result of ``equivalence'' is equal to the multiplication of real number. For convenience, from here we will treat the binary values as real numbers and use the corresponding real number operation instead of the Boolean operation (just like the  ``equivalence'' to multiplication). As shown in Fig.~\ref{ANNbasic}, assume that $\bm{\vec x}=(x_1,x_2,...,x_n)$ are the input neurons of BDNN and $\bm{\vec w}=(w_1,w_2,...,w_n)$ are the corresponding weights, $y$ is the output neuron, then we have $x_i,w_i,y \in \{-1,1\}$. As mentioned above, the linear calculation of BDNN is 
\begin{equation*}
\bm{\vec x} \times \bm{\vec w}^T = (x_1 w_1, x_2 w_2,...,x_n w_n).
\end{equation*}
\par

\begin{table}[t]
\caption{Truth table of Boolean operations.\label{truthtable}}
\centering
\begin{tabular}{c c!{\vrule width 1.5pt}c|c|c|c|c}
$a$ & $b$ & $a\wedge b$ & $a\vee b$ & $a\oplus b$ & $a\equiv b$ & $a \rightarrow b$ \\ \hline
$0$ & $0$ & $0$         & $0$       & $0$         & $1$         & $1$               \\
$1$ & $0$ & $0$         & $1$       & $1$         & $0$         & $0$               \\
$0$ & $1$ & $0$         & $1$       & $1$         & $0$         & $1$               \\
$1$ & $1$ & $1$         & $1$       & $0$         & $1$         & $1$               \\
\end{tabular}
\end{table}

Second, the nonlinear calculation of BDNN is defined as follows: we count the number of ``$-1$'' and ``$1$'' in $(x_1 w_1, x_2 w_2,...,x_n w_n)$ and let the calculation return the one with larger number. Assume $f$ is the basic function of BDNN, with the linear and nonlinear calculation defined above, then we have
\begin{equation}
y=f(\bm{\vec x},\bm{\vec w}) = \begin{cases} 1,& if \quad \sum_{i=1}^{n}x_i w_i \geq 0 \\ -1,& if \quad \sum_{i=1}^{n}x_i w_i < 0.
\label{f}
\end{cases}
\end{equation}
\par
\begin{figure}[h]
\begin{center}
\includegraphics[width=0.3\textwidth]{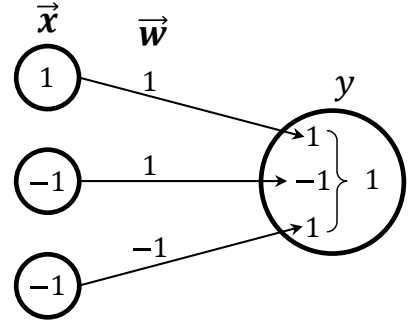}
\caption{The calculation of BDNN basic function $f$. \label{BDNNexample}}
\end{center}
\end{figure}

Figure~\ref{BDNNexample} shows an example of the calculation of \eqref{f}. The inputs $\bm{\vec x}$ are $(1,-1,-1)$ and the corresponding weights $\bm{\vec w}$ are $(1,1,-1)$. Then, we use Boolean operation as linear calculation and the results $(1,-1,1)$ are obtained. Finally, since in the results, there are two ``$1$'' and one ``$-1$'', thus the value of output $y$ is ``$1$''. Please note that although \eqref{f} is written as real number function, it denotes the Boolean operation of ``equivalence'' and a counting operation. With \eqref{f}, the forward propagation of BDNN can be realized. For a certain set of input binary data, the BDNN can calculate its corresponding output, which is also binary. 
\par   

\subsection{The hybrid-BDNN}

Clearly, the BDNN introduced above is only able to process binary input data. If the input data is non-binary, such as grayscale image, the BDNN can not be used directly. Therefore, we should first convert the non-binary data into binary before the BDNN is used. For such situation, we proposed the hybrid-BDNN, which is a combination of conventional neural network and BDNN.\par

\begin{figure*}[t]
\begin{center}
\includegraphics[width=0.7\textwidth]{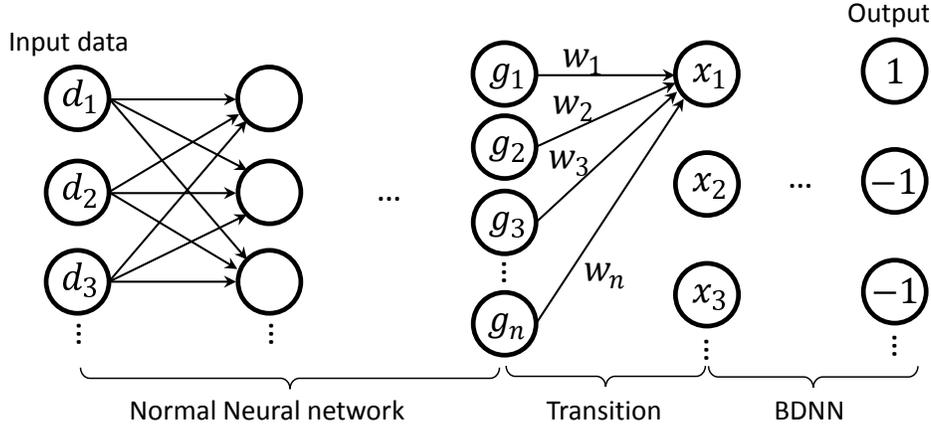}
\caption{The structure of hyrbid-BDNN. \label{HybridBDNN}}
\end{center}
\end{figure*}

As shown in Fig.~\ref{HybridBDNN}, the Hybrid-BDNN contains three parts: the normal neural network part, the transition part and the BDNN part. The lower layers of hybrid-BDNN are normal neural network part, which is connected to the input data (binary or non-binary); the higher layers are the BDNN part, which generates the result. The transition part is a single layer between the normal neural network part and BDNN part, by which the two different neural networks are combined.\par

The basic function of the normal neural network part is the same with the conventional networks (inner product and activation function); the basic funtion of BDNN is just introduced above. Consequently, the forward propagation can be conducted in both the normal neural network and the BDNN part. The remaining problem is the transition part. If we define the basic function of the transition part, then we can conduct the forward propagation of the whole hybrid-BDNN. \par   

As shown in Fig.\ref{HybridBDNN}, assume that $\bm{\vec g}=(g_1,g_2,...,g_n)$ are the input neurons of the transition part and the $(x_1,x_2,x_3...)$ are the output neurons. As mentioned above, $\bm{\vec g}$ belongs to the normal neural network part, thus it is all real number. The $(x_1,x_2,x_3...)$ belongs to the BDNN part, so they are all binary values. Let $t$ denote the basic function of transition part, if we take $x_1$ as example and $\bm{\vec w}=(w_1,w_2,...,w_n)$ are the corresponding weights, then we obtain 
\begin{equation*}
t(\bm{\vec g},\bm{\vec w}) = x_1.
\end{equation*}
Obviously, in order to calculate with $\bm{\vec g}$, the $\bm{\vec w}$ is also real number. Then we define $t$ as follows: 
\begin{equation}
x_1=t(\bm{\vec g},\bm{\vec w}) = \begin{cases} 1,& if \quad A(\sum_{i=1}^{n}g_i w_i) \geq T \\ -1,& if \quad A(\sum_{i=1}^{n}g_i w_i) < T,
\label{t}
\end{cases}
\end{equation}
where $A$ is activation function and $T$ is a fixed threshold. In fact, the calculation $A(\sum_{i=1}^{n}g_i w_i)$ is just the basic function of the normal neural network part. The \eqref{t} means that we use a threshold $T$ to convert the output real number of the normal neural network part into binary value.
\par

With \eqref{t}, the forward propagation of the whole hybrid-BDNN can be realized. This network can be used for any input data. By the way, if the range of activation function $A$ is $(-1,1)$, then we can set the threshold $T=0$. This is convenient for the training.  
\par

In summary, the BDNN is very suitable for binary input data classification, such as binary image of characters. If the input data is non-binary, the hybrid-BDNN can be used. Please note that it is better not to use too many layers in normal neural network part, otherwise the computation speed of hybrid-BDNN may be slowed much. In the  experiments, we only used one layer (including the input data) as the normal neural network part.
\par

\section {Training method for BDNN and hybrid-BDNN} \label{training}

Usually, the gradient descent algorithm is seen as the training method for neural networks. For the training of BDNN and hybrid-BDNN, we also used this algorithm. However, compared with the conventional neural network, the BDNN has obviously different properties, thus we can not directly use this algorithm on BDNN training. In order to solve this problem, some approximation and conversion are applied to BDNN to make it suitable for gradient descent training.
\par

\subsection{Gradient descent training for BDNN}  

In the conventional gradient descent training, in each iteration, the weights are adjusted by a small value (depends on the error propagation and learning rate). Nevertheless, since the weights in BDNN are binarized, thus it is hard to adjust the weights in the same way. Hence, in order to apply the gradient descent algorithm, in the training of BDNN, the real numbers are used instead of the binary values. A conversion function $C$ is defined to convert the real numbers to corresponding binary values, which is   
\begin{equation}
C(x) = \begin{cases} 1,& if \quad x \geq 0 \\ -1,& if \quad x < 0.
\label{C}
\end{cases}
\end{equation}
With function \eqref{C}, we can convert the trained weights of real number to binary value.
\par 

\begin{figure*}[t]
\begin{center}
\includegraphics[width=0.85\textwidth]{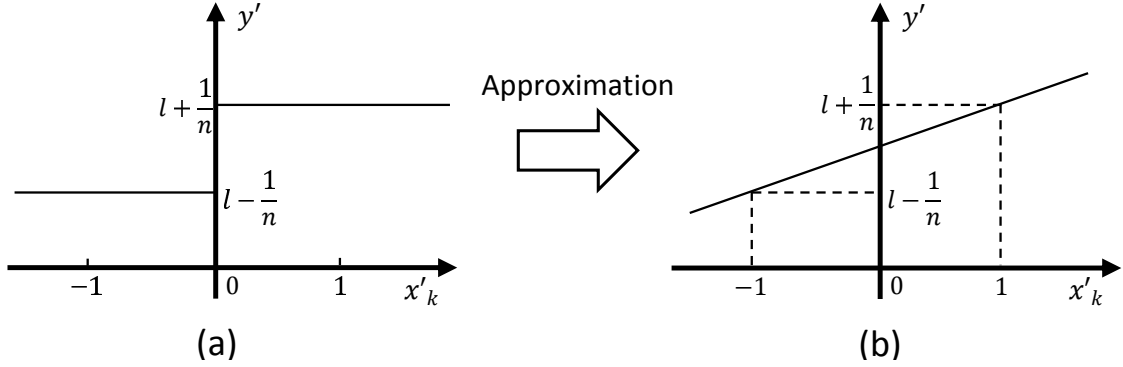}
\caption{The curve of output neuron by a certain input neuron and its approximation. \label{curve}}
\end{center}
\end{figure*}

However, the key point of using real number is that we must keep the forward propagation result the same with the binarized network; otherwise, the training is meaningless. Therefore, in order to satisfy this request, a basic function $f'$ for training is defined. First, assume $\bm{\vec x}=(x_1,x_2,...,x_n)$, $\bm{\vec w}=(w_1,w_2,...,w_n)$ and $y$ are the input neurons, weights and output of the BDNN; $\bm{\vec x'}=(x_1',x_2',...,x_n')$, $\bm{\vec w'}=(w_1',w_2',...,w_n')$ and $y'$ are the corresponding real numbers of training. Second, assume
\begin{equation}
C(x_i') = x_i,\quad C(w_i') = w_i.
\label{CC}
\end{equation}
Then, the basic function $f'$ of training is given by 
\begin{equation}
y' = f'(\bm{\vec x'}, \bm{\vec w'}) = \dfrac{ \sum_{i=1}^{n}\dfrac{x_i' w_i'}{\left|x_i' w_i'\right|} }{ n }.
\label{fp}
\end{equation}
As mentioned above, in \eqref{f}, the basic function $f$ of BDNN counts the number of ``1'' and ``-1'' in 
$(x_1 w_1, x_2 w_2,...,x_n w_n)$ and returns the one with larger number. Accordingly, in \eqref{fp}, for the basic function $f'$, if in $(x_1' w_1', x_2' w_2',...,x_n' w_n')$ the positive values are more than negative ones, then $y'$ is a positive value; otherwise, $y'$ is negative. Consequently, because of \eqref{CC}, we obtain
\begin{equation*}
C( f(\bm{\vec x'},\bm{\vec w'}) ) = f(\bm{\vec x},\bm{\vec w})
\end{equation*}
and
\begin{equation*}
C( y' ) = y.
\end{equation*}
\par

Clearly, with \eqref{fp}, if the inputs of training are the same with BDNN, then after forward propagation, each real number neuron of training equals to the corresponding binary neuron of BDNN (by using conversion function $C$), including the output neurons. As a result, we can train the BDNN with real numbers. After training, we just need to convert the real number weights into binary values and then a trained BDNN is obtained. 
\par 

After the forward propagation of training is solved, the next problem is the backpropagation. In order to use gradient descent algorithm, we must define two partial derivatives: the $\dfrac{\partial y'}{\partial x_i'}$ and $\dfrac{\partial y'}{\partial w_i'}$. With such two partial derivatives, the backpropagation can be conducted. 
\par

Assume $x_k'$ is a certain neuron from $\bm{\vec x'}$ and $w_k'$ (let $w_k'>0$) is its corresponding weight. Let $l$ take the form
\begin{equation*}
l = \dfrac{ \sum_{i=1, i\neq k}^{n}\dfrac{x_i' w_i'}{\left|x_i' w_i'\right|} }{ n }.
\end{equation*}
Then, we have
\begin{equation*}
y' = l+\dfrac{ x_k' w_k' }{ n\left|x_k' w_k'\right| }.
\end{equation*}
It is obvious that $l$ and $w_k'$ are independent. Thus we can draw a curve of $y'$ by $w_k'$, which is shown in Fig.~\ref{curve} (a). This curve is not continuous. We can see that the $\dfrac{\partial y'}{\partial x_k'}$ is $+\infty$ at $x_k'=0$ and $0$ at the rest of the positions. Clearly, we can not use such partial derivative for gradient descent algorithm. Therefore, an approximation of this curve (shown in Fig.~\ref{curve} (b)) is used to calculate the $\dfrac{\partial y'}{\partial x_k'}$. The new curve is a strait line, go through the point $(-1, l-\dfrac{1}{n})$ and $(1, l+\dfrac{1}{n})$. Here, we assume that $w_k'>0$. If $w_k'<0$, the slope of the approximation line will be reversed. Consequently, by using the approximation, we obtain
\begin{equation}
\dfrac{\partial y'}{\partial x_k'}=\dfrac{2w_k'}{n\left|w_k'\right|}.
\label{dx}
\end{equation}
Similarly, since in function $f'$, the $x_k'$ and $w_k'$ are symmetric, so that
\begin{equation}
\dfrac{\partial y'}{\partial w_k'}=\dfrac{2x_k'}{n\left|x_k'\right|}.
\label{dw}
\end{equation}
Here, in \eqref{dx} and \eqref{dw}, the $\frac{ 2 }{ n }$ is the value of the slope; the $\dfrac{w_k'}{\left|w_k'\right|}$ or $\dfrac{x_k'}{\left|x_k'\right|}$ determines the sign of the slope. Please note that the slope value can be set to not only $\frac{2}{n}$ but also other proper values.
\par

With \eqref{dx} and \eqref{dw}, the gradient descent algorithm can be applied to train the BDNN. Although an approximation was used, the following experimental results showed that, the BDNN could be trained well with such method.
\par

\subsection{The training of transition part of hybrid-BDNN} 

As shown in Fig.~\ref{HybridBDNN}, now the backpropagation can be applied to the normal neural network part and the BDNN part. However, in order to apply the backpropagation on hybrid-BDNN, we still need to consider about the backpropagation of the transition part. 
\par

First, the same with the BDNN training, in the training of the transition part, the real numbers are used instead of the binary values. Assume that $x_1'$ is the corresponding real number of $x_1$ and function $t'$ is the real number version of function $t$. Then, the same with $f'$, in order to keep the same forward propagation result, $t'$ should satisfy the following equation
\begin{equation*}
C( t'(\bm{\vec g},\bm{\vec w}) ) = C(\bm{\vec g},\bm{\vec w}).
\end{equation*}
Therefore, we define $t'$ as
\begin{equation}
t'(\bm{\vec g}, \bm{\vec w})=A(\sum_{i=1}^{n}g_i w_i)-T.
\label{tp}
\end{equation}
Obviously, the $t'$ of \eqref{tp} satisfies the request. If we set $T=0$, then $t'$ is just the activation function. Naturally, for a certain neuron $g_k$ and its weight $w_k$, the two derivatives of the transition part are given by
\begin{equation}
\dfrac{\partial x_1'}{\partial g_k}=\dfrac{\partial A}{\partial g_k}, \quad \dfrac{\partial x_1'}{\partial w_k}=\dfrac{\partial A}{\partial w_k}.
\label{dt}
\end{equation}
In \eqref{dt}, the two partial derivatives are just the same with the normal neural network. With \eqref{dt}, the backpropagation of the transition part becomes possible.   
\par

Consequently, by using the above method, the backpropagation can be applied to the whole hybrid-BDNN. Now we can train the hybrid-BDNN by using the gradient descent algorithm.
\par

\subsection{Special training technique for BDNN}

Clearly, the training method for BDNN proposed in this paper is very close to the conventional training method. Thus it is possible to borrow the training techniques of conventional neural networks to BDNN, such as dropout and dropconnect. Besides, there are several special training techniques, which are necessary for BDNN.
\par

First, in the training, we must keep the $n$ of \eqref{fp} an odd number all the time. This is because the output $0$ of $f'$ is not defined. The result of $f'$ should be either a positive number or a negative number. Consequently, in the following experiments, in the BDNN, the neuron number $n$ was always an odd number.
\par

Second, in the training of conventional neural network, the error of the output is determined by the difference between the output neuron and its ground truth. However, in the training of BDNN, the error is determined by not only the difference but also the sign of each value. Assume the error of a certain output neuron $y$ is $e$ and the corresponding ground truth is $y^T$, then the error calculation is given by
\begin{equation}
e = \begin{cases} 0,& if \quad yy^T > 0 \\ \frac{1}{2}(y-y^T),& if \quad yy^T < 0.
\end{cases}
\label{e}
\end{equation}
The\eqref{e} means that if the output neuron and its ground truth are both positive or negative, then the error of this output is $0$; otherwise, the error is calculated as conventional neural network. Actually, according to \eqref{C}, if the output neuron has the same sign with its ground truth, then the output is seen as correct. This technique is essential for the BDNN training, without which the training can not converge.
\par

\section{Experiments}
\label{experiments}

In the experiments, two different network structures were studied --- the classical three-layer network and the convolutional neural network (CNN). Besides the BDNN and hybrid-BDNN, the conventional neural networks of the same structures were also tested for comparison. Since now we only have the basic training method for BDNN, so in order to make fair comparison, most of the training optimization techniques of the conventional neural network were not used. For example, in the experiments, the learning rate was fixed and the simple hyperbolic tangent function was used as activation function.  
\par

Moreover, two datasets, MNIST and CIFAR10 were used for the experiments. As we know, MNIST is a dataset of handwritten digits, thus it is suitable for binary data classification test. In contrast, the CIFAR10 was used as non-binary data, on which the hybrid-BDNN was tested. In each dataset, the training data was used to train the networks; the test data was used as validation data. After the training was finished, the iteration with the lowest error rate on test data was seen as the final result. 
\par

\subsection{The classical three-layer network}

\begin{figure}[t]
\begin{center}
\includegraphics[width=0.3\textwidth]{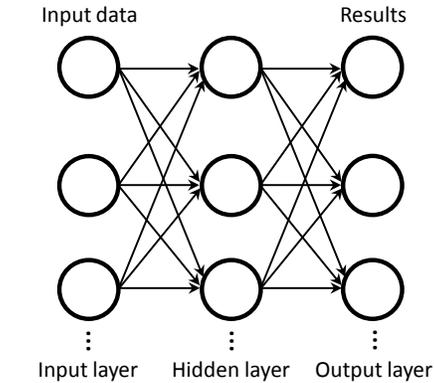}
\caption{The structure of three-layer network. \label{3layer}}
\end{center}
\end{figure}

In the early stages of ANN, before the DNN was introduced, the three-layer network was widely used for classification tasks. As shown in Fig.~\ref{3layer}, in such network, there are three different layers: the input layer, the hidden layer and the output layer. The input layer contains the input data while the output layer generates the classification results. Besides, all the layers are fully connected. Since this network structure was very classic, so the BDNN of this structure was first tested.
\par

\begin{figure*}[t]
\begin{center}
\includegraphics[width=0.85\textwidth]{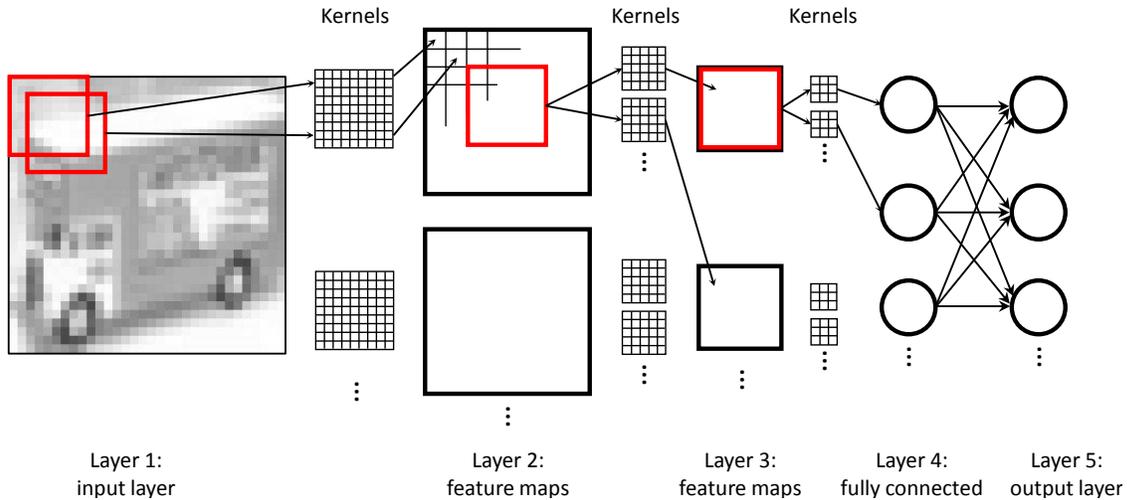}
\caption{The network structure of CNN. \label{cnn}}
\end{center}
\end{figure*}

For the experiments of three-layer BDNN, the binary data of MNIST was used. Because the original images of MNIST were grayscale, so we used a simple binarization method (fixed threshold) to convert the grayscale images into binary. All the training data of MNIST was used to train the network and the test data of MNIST was used as validation data.
\par

In MNIST, the image size is $28\times28$, thus there are 784 pixels, which are used as input neurons. As mentioned above, in order to make the neurons of the input layer an odd number, one neuron was simply added to the input layer and set to $1$. Consequently, the input layer contains 785 neurons. Moreover, since the digits have 10 classes, so the output layer is set to 10 neurons. 
\par

For the hidden layer, we used two different numbers, one is 1571 (double of the input neuron number) and the other is 2355 (three times of the input neuron number). Then, two different network structures are obtained as follows.
\begin{itemize}
\item Structure A: the input layer contains 785 neurons, followed by a hidden layer of 1571 neurons, then an output layer of 10 neurons (one neuron stands for one digit class). 
\item Structure B: the input layer contains 785 neurons, followed by a hidden layer of 2355 neurons, then an output layer of 10 neurons (one neuron stands for one digit class).
\end{itemize} 
In the experiments, the BDNN of both structures were tested. Meanwhile, the conventional ANN of Structure A was also tested for comparison. 
\par

\begin{table}[t]
\caption{Error rates (\%) of the three-layer networks on MNIST. \label{3layerresult}}
\centering
\begin{tabular}{c !{\vrule width 1.5pt} c c|c c}
          	&\multicolumn{2}{c|}{\textbf{BDNN}}	& \multicolumn{2}{c}{\textbf{Normal ANN}} 	\\
			& \emph{Train} & \emph{Test} 	& \emph{Train} & \emph{Test} 	\\ \hline
Struct. A 	& 20.73	& 16.74	& 1.07	& 1.64	\\
Struct. B 	& 16.89	& 14.50	&		& 		\\
\end{tabular}
\end{table}

The experimental results are shown in Table~\ref{3layerresult}. On one hand, by using the proposed training method, finally, the BDNN converged on the training dataset and on the test dataset, the classification rate was around $15\%$. By this result, the feasibility of BDNN is proved. On the other hand, it can be seen that the conventional ANN of Structure A has much lower error rate than the BDNN of same structure. This is may be because the BDNN was not fully trained by the proposed method (borrowed from the conventional network training). If we can design specific training method for BDNN, the performance of which may be improved a lot. Another possible reason may be the complexity of the network. Since the binary neurons are used in BDNN, the complexity of which is lower than the conventional ANN, though they have the same structure. The result of BDNN of Structure B indicates that, when the scale of BDNN is increased (higher complexity), the performance becomes better.
\par

\subsection{The convolutional neural network}

In deep learning research, the structure of CNN is widely used for image classification. Therefore, in the experiments, we also tested the CNN structure. As shown in Fig.~\ref{cnn}, we used a CNN structure of five layers. The first layer is the input data, followed by two layers of feature maps (calculated by the convolutional kernels). The last two layers are the fully connected neurons and one of them is used as output layer. In most of the CNNs, after convolutional operation, the feature map size is then decreased by pooling. Nevertheless, since it is hard to realize the pooling operation on binary values, thus the feature map size is controlled by skipping every other pixel in the convolutional operation. 
\par

With different sizes of the feature maps and kernels, three different CNN structures are obtained as follows.     
\begin{itemize}
\item Structure C: Layer 1 is the input image, the size is $29 \times 29$; Layer 2 contains $7$ feature maps of size $13 \times 13$, obtained by using $7$ kernels of size $5 \times 5$; Layer 3 contains $51$ feature maps of size $5 \times 5$, obtained by using $357$ kernels of size $5 \times 5$; Layer 4 contains 201 neurons, obtained by using $10,251$ kernels of size $5 \times 5$; Layer 5 contains 10 neurons for output. 
\item Structure D: Layer 1 is the input image, the size is $29 \times 29$; Layer 2 contains $17$ feature maps of size $13 \times 13$, obtained by using $17$ kernels of size $5 \times 5$; Layer 3 contains $71$ feature maps of size $5 \times 5$, obtained by using $1,207$ kernels of size $5 \times 5$; Layer 4 contains 201 neurons, obtained by using $14,271$ kernels of size $5 \times 5$; Layer 5 contains 10 neurons for output.
\item Structure E: Layer 1 is the input image, the size is $33 \times 33$; Layer 2 contains $17$ feature maps of size $5 \times 5$, obtained by using $17$ kernels of size $9 \times 9$; Layer 3 contains $71$ feature maps of size $5 \times 5$, obtained by using $1,207$ kernels of size $5 \times 5$; Layer 4 contains 201 neurons, obtained by using $14,271$ kernels of size $5 \times 5$; Layer 5 contains 10 neurons for output.
\end{itemize} 
The Structure C and D were used by BDNN and tested on MNIST. Structure E was used by the hybrid-BDNN and tested on CIFAR10 (non-binary data). Therefore, for Structure E of hybrid-BDNN, the input layer and the following kernels were real numbers (normal neural network part) and the rest of the network was the BDNN part.
\par

First, the error rates of MNIST are shown in Table~\ref{CNNIST}. Clearly, by using deeper structure (more layers), the performance of BDNN was improved. Meanwhile, the normal CNN of the same structure still had much better result. In Structure D, when more feature maps were used, the result of BDNN became better. This also shows the potential of BDNN of larger scales.
\par

\begin{table}[t]
\caption{Error rates (\%) of the CNN structures on MNIST. \label{CNNIST}}
\centering
\begin{tabular}{c !{\vrule width 1.5pt} c c|c c}
          	&\multicolumn{2}{c|}{\textbf{BDNN}}	& \multicolumn{2}{c}{\textbf{Normal CNN}} 	\\
			& \emph{Train} & \emph{Test} 	& \emph{Train} & \emph{Test} 	\\ \hline
Struct. C 	& 11.78	& 9.98	& 0.26	& 1.10	\\
Struct. D 	& 7.11	& 6.54	&		& 		\\
\end{tabular}
\end{table}

Second, the error rates of CIFAR10 are shown in Table~\ref{CNNCIFAR}. CIFAR10 is a database of color images of 10 different kinds of objects. In the experiment, the color images were converted into grayscale and then used. Since this database was very difficult, the error rates of both networks were very high, though the largest network structure were used.
\par

Another observation is that although the training error rate of hybrid-BDNN was much lower than CNN, its test error rate was lower. This indicates that (i) it is possible to use hybrid-BDNN for non-binary data classification and (ii) larger network structure of high complexity is necessary for reducing the training error of hybrid-BDNN. 
\par    

\begin{table}[t]
\caption{Error rates (\%) of the CNN structures on CIFAR10. \label{CNNCIFAR}}
\centering
\begin{tabular}{c !{\vrule width 1.5pt} c c|c c}
          	&\multicolumn{2}{c|}{\textbf{Hybrid-BDNN}}	& \multicolumn{2}{c}{\textbf{Normal CNN}} 	\\
			& \emph{Train} & \emph{Test} 	& \emph{Train} & \emph{Test} 	\\ \hline
Struct. E 	& 89.84	& 87.48	& 	41.71	& 	88.37	\\

\end{tabular}
\end{table}

\section{Conclusion and future work}
\label{conclusion}

In BDNN, we first tried the brand new binarized propagation function for neural networks. Moreover, a special gradient descent training method is also proposed for BDNN. The experimental results proved that it is able to use BDNN just like the conventional DNN. Besides changing the network structure and training strategy, our trial may provide new thought of improving the DNN and extend its usage.
\par

Because of the computational limitation, we didn't test the BDNN of large scale in this paper. Therefore, the performance of BDNN may not be comparable with the state-of-the-art results. In the future, we will develop GPU based training program to train the BDNN of larger scale. Various network structures will also be studied, such as the deep fully connected neural network, the recurrent neural network and so on. Moreover, the optimization of the forward propagation of the BDNN on CPU will be studied.   
\par

\section{Acknowledgments}

This work was started at the beginning of 2014 and based on this work, a patent (China) is applied in November, 2014. The application number of the patent is 201410647710.3. We also submitted this paper to CVPR 2015 but it was rejected. However, we still believe this work is valuable and the BDNN is a promising solution for low-performance device to use deep learning models.\par 

{\small
\bibliographystyle{ieee}
\bibliography{egbib}
}

\end{document}